\documentclass[journal]{IEEEtran}
\usepackage[cmex10]{amsmath}
\usepackage{bm}
\usepackage[switch]{lineno}
\usepackage{caption}
\usepackage{amsfonts}
\usepackage{booktabs}
\usepackage{comment}
\usepackage{subfig}
\usepackage{graphicx}
\usepackage{tabu}
\ifCLASSINFOpdf
\else
\fi
\begin{document}
%
\title{PH-GCN: Person Re-identification with Part-based Hierarchical Graph Convolutional Network}
%
%
%

\author{Bo~Jiang,
        Xixi~Wang,
        and Bin~Luo
\IEEEcompsocitemizethanks{\IEEEcompsocthanksitem Bo Jiang, Xixi Wang and Bin Luo are with the School of Computer Science and Technology, Anhui University, Hefei, China \protect\\
E-mail: jiangbo@ahu.edu.cn}}
\maketitle

\begin{abstract}
The person re-identification (Re-ID) task requires to robustly extract feature representations for person images.
Recently, part-based representation models have been widely studied for extracting the more compact and robust feature representations for person images to improve person Re-ID results.
However, existing part-based representation models mostly extract the features of different
 parts independently which ignore the relationship information between different parts.
 To overcome this limitation, in this paper we propose a novel
 deep learning framework, named  Part-based Hierarchical Graph Convolutional Network (PH-GCN) for person Re-ID problem.
 Given a person image, PH-GCN first constructs a hierarchical graph to represent the pairwise relationships among different parts.
Then, both local and global feature learning are performed by the messages passing in
PH-GCN, which takes other nodes information into account for part feature representation.
Finally, a perceptron layer is adopted for the final person part label prediction and re-identification.
The proposed framework provides a general solution that integrates \emph{local}, \emph{global} and \emph{structural} feature learning simultaneously in a unified end-to-end network.
Extensive experiments on several benchmark datasets demonstrate the effectiveness of the proposed PH-GCN based Re-ID approach.
\end{abstract}


%
\IEEEpeerreviewmaketitle

\section{Introduction}
%
%
%
%
Person re-identification (Re-ID) is an active research problem in computer vision~\cite{Cheng2016Person,si2018dual,Sun2017SVDNet,shen2018deep,chen2018group,li2018harmonious,Chen_2018_CVPR,ustinova2017multi}.
Many of existing Re-ID methods adopt a person classification framework which aims to
determine the label of an input person image by using a classifier trained on the training samples~\cite{Cheng2016Person,su2017pose,schumann2017person,si2018dual,chen2018group,sun2018beyond}.
Although recent years have witnessed rapid advancements in person Re-ID, it is still a challenging task partly due to large changes of person appearance caused by variety of factors, such as pose, illumination, deformation and occlusion.

One main issue for Re-ID problem is to develop a compact and robust feature representation for person images.
Recently, part-based methods have been widely studied and verified beneficially to person Re-ID task~\cite{Zhao_2017_ICCV,su2017pose,suh2018part,sun2018beyond,Cheng2016Person,wei2017glad,zheng2018coarse}.  These methods
generally conduct feature representation on part-level  and thus can extract both local and global discriminative representations for person image.
In particular,
deeply-learned features have been verified stronger discriminative ability, especially when aggregated from
deeply-learned part features~\cite{su2017pose,Zhao_2017_ICCV,sun2018beyond,zheng2018coarse,wang2018learning}.
For example,
Zhao $et\ al.$ \cite{Zhao_2017_ICCV} develop a human part-aligned representation for Re-ID. 
Sun $et\ al.$ \cite{sun2018beyond} propose Part-based Convolutional Baseline (PCB) for learning part-level features.
%
Wei $et\ al.$ \cite{wei2017glad} propose Global-Local-Alignment descriptor (GLAD) to leverage both local and global cues. 
Zheng $et\ al.$ \cite{zheng2018coarse} design a new coarse-fine pyramid model to conduct local and global representation. 

%
%

\textbf{Motivation.} The above existing person Re-ID models mostly extract the features of different
person parts independently which ignore the inherent spatial relationship information among different parts.
Such spatial relationships are usually represented in the form of graphs.
Recently, graph convolutional networks (GCNs) have draw increasing attention due to their ability of generalizing neural networks for data with graph
structures~\cite{bruna2013spectral,kipf2016semi,niepert2016learning,defferrard2016convolutional,yan2018spatial}.
GCNs aim to propagate messages on a graph structure. After message passing on the graph, the final node representations are obtained from their own as well as the information of their neighboring nodes, which thus can naturally incorporate the contextual information for node representations.
Also, GCNs incorporate graph computation into the neural networks learning to make the training end-to-end, which can be naturally incorporated into some other deep learning frameworks.
Motivated by these, in this paper we propose a novel Part-based Hierarchical Graph Convolutional Network (PH-GCN)  for person image representation and Re-ID task.
PH-GCN aims to learn a context-aware representation for each person part that incorporates the geometrical structure information among parts while maintains the unary appearance feature of each part.
PH-GCN exploits the inherent relationships of parts effectively, and thus is robust to part noises and/or corruptions.

Overall, the main contributions of this paper are summarized as follows.
\begin{itemize}
 \item We propose a novel deeply-learned and context-aware part feature extraction and learning model
 for person Re-ID.  We show that the proposed representation has much higher discriminative
ability than  traditional part-based feature representations.
  \item We propose a novel Part-based Hierarchical Graph Convolutional Network (PH-GCN) learning framework for  object representation.
      The proposed framework provides a general solution that integrates \textbf{local}, \textbf{global} and \textbf{structural} feature representation and learning simultaneously in a unified network.
\end{itemize}
Extensive experiments on several benchmark datasets demonstrate the effectiveness of the proposed PH-GCN method.

\section{Related Works}

\subsection{Person re-identification}

Many methods have been proposed for person Re-ID problem~\cite{Cheng2016Person,schumann2017person,si2018dual,shen2018deep,li2018harmonious,Chen_2018_CVPR,ustinova2017multi}.
In this section, we briefly review some recent related works that are also devoted to generate deeply-learned part features for Re-ID problem.

Ustinova $et\ al.$ \cite{ustinova2017multi} propose a network architecture to learn a more effective embedding by performing bilinear pooling.
Si $et\ al.$ \cite{si2018dual} propose to develop Dual Attention Matching network (DuATM) to learn context-aware feature sequences. 
Su $et\ al.$ \cite{su2017pose} propose Pose-driven Deep Convolutional (PDC) model, which aims to utilize the human part cues to alleviate the pose variations and thus learn robust features from both global image and local parts.
Zhao $et\ al.$ \cite{Zhao_2017_ICCV} propose a human part-aligned representation by detecting the human body regions and computing  between the corresponding parts.
Sun $et\ al.$ \cite{sun2018beyond} propose a strong convolution baseline method to further leverage a uniform partition strategy and thus learn a more compact representation for person Re-ID.
Chen $et\ al.$ \cite{chen2018group} propose to combine CRF and deep neural network together and use multiple images to model the relationships between local similarity and global similarity. 
To further incorporate the global information,
%
Wei $et\ al.$ \cite{wei2017glad} propose Global-Local-Alignment Descriptor (GLAD) that explicitly leverages the local and global cues in human body to generate a discriminative and robust representation.
Wang $et\ al.$ \cite{wang2018learning} develop a feature learning strategy to integrate discriminative information by combining global and local information in different granularities.
Zheng $et\ al.$ \cite{zheng2018coarse} propose a new coarse-fine pyramid model to conduct local and global representation simultaneously. 

\subsection{Graph convolutional network}

Recently, graph convolutional neural networks (GCNs) have been demonstrated effectively for graph structure data representation and learning in machine learning area~\cite{bruna2013spectral,kipf2016semi,niepert2016learning,defferrard2016convolutional,yan2018spatial}.
For example,
Bruna $et\ al.$ \cite{bruna2014spectral} propose a CNN-like neural architecture
on graphs in Fourier domain.
Defferrard $et\ al.$ \cite{defferrard2016convolutional} propose to approximate the spectral filters by using recurrent Chebyshev polynomials. 
Kipf and Welling \cite{kipf2016semi} further propose a simplified Graph Convolutional Network (GCN) based on the first-order approximation of spectral filters.
Atwood and Towsley \cite{atwood2016diffusion} propose Diffusion-Convolutional Neural Networks (DCNNs). 
Monti $et\ al.$ \cite{monti2017geometric} present mixture model CNNs (MoNet) to generalize CNN architecture on graphs.
Veli{\v c}kovi{\'c} $et\ al.$ \cite{velickovic2017graph} present Graph Attention Networks (GAT) by further designing an edge attention layer. 

In addition, GCNs (or GNNs) have also been employed in computer vision tasks~\cite{guo2018neural,yan2018spatial,qi20173d,shen2018person}.
For example,
Michelle $et\ al.$ \cite{guo2018neural} develop neural graph matching networks for few-shot 3D action recognition.
Yan $et\ al.$ \cite{yan2018spatial} propose  Spatial Temporal Graph Convolutional Network (ST-GCN) for skeleton-based action recognition.
Qi $et\ al.$ \cite{qi20173d} propose a 3D graph neural network model for rgbd semantic segmentation.
Shen $et\ al.$ \cite{shen2018person} propose            Deep Similarity-Guided Graph Neural Network (SGGNN) to explore the geometric relationships among different person images for person Re-ID.
Different from SGGNN \cite{shen2018person}, here we exploit GCN for person feature representation.
To the best of our knowledge, we
are the first to employ GCN model to extract the context-aware part-based representation for person Re-ID task.

\begin{figure*}[ht]
\centering
\noindent\makebox[\textwidth][l] {
\includegraphics[height=5cm,width=17.5cm]{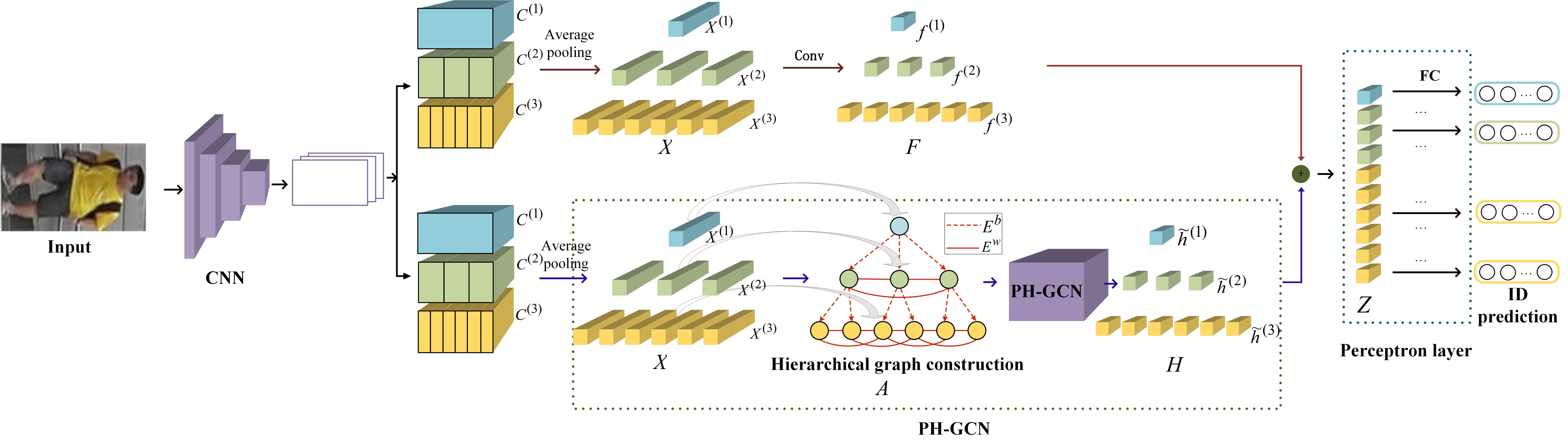}
}
\caption{Architecture of the proposed PH-GCN network for person Re-ID, which contains CNN based part feature extraction and hierarchical graph construction, graph convolutional module and perceptron layer.}
\end{figure*}

\section{PH-GCN Model}

In this section, we propose our Part-based Hierarchical Graph Convolutional Network (PH-GCN) for person image representation and re-identification.

\subsection{Overview}

Figure 1 shows the overall network module of PH-GCN which contains CNN based part feature extraction and part-based hierarchical graph construction, graph convolutional module and perceptron layer.

\begin{itemize}
  \item  {\bf CNN based part feature extraction:} We utilize a deep CNN network module to extract the appearance feature for each part of person image.
  \item  {\bf Part-based hierarchical graph construction:} A hierarchical graph is constructed to encode the spatial relationships among different person parts. 
  \item  {\bf Graph convolutional module:} We employ a graph convolutional network (GCN) architecture to extract the context-aware representations for person parts. 
  \item {\bf Perceptron layer:} A perceptron layer is defined for final Re-ID task. 
\end{itemize}
In the following, we present the details of each module in our network, respectively.
%

%

\subsection{CNN based part feature extraction}

For each person image ${I}$, we first use a pre-trained CNN model to extract convolutional feature descriptor for it.
Then, we conduct three uniform partitions $\mathcal{C}^{(p)}, p=1, 2, 3$ on the conv-layer respectively to obtain global and local part-level features, as shown in Figure 1.
In particular, we adopt ResNet50~\cite{He_2016_CVPR} network and conduct partition on the 4th conv-layer. 
The parameters of ResNet50 network are pretrained on ImageNet~\cite{russakovsky2014imagenet}.
Finally, we  obtain a feature descriptor $x^{p}_i$ for the $i$-th part in partition $\mathcal{C}^{(p)}$ by using an average pooling strategy. 
We denote $X^{(p)} = (x^{p}_1, x^{p}_2 \cdots x^{p}_{n_p}) \in \mathbb{R}^{d\times n_p}, p=1,2,3$ as the feature collections for different parts of image $I$ in the following sections.

\subsection{Hierarchical part graph construction}

Based on the above hierarchical partitions $\mathcal{C}^{(p)}, p=1, 2, 3$,
we construct a hierarchical graph $G = (V, E)$  to represent the spatial and appearance  relationships among different parts.
In particular, we construct a three-layer graph whose nodes and edges are introduced below.
%
%
%
%

 {\textbf{Nodes.} }
A three-layer hierarchical graph $G = (V, E)$ is constructed, where $V=\{V^{(1)},V^{(2)}, V^{(3)}\}$ with each $V^{(p)}$  corresponding to the partition level $\mathcal{C}^{(p)}, p=1, 2, 3$.
Each node $v^{(p)}_i \in V^{(p)}$ corresponds to a specific part which is assigned with a feature vector $x^p_i$, and there exist more nodes on the higher layers, i.e., $|V_3|> |V_2|>|V_1|$.
The higher layer contains more local information while the lower layer represents the global representation of person.
We set $|V_3|, |V_2|$ and $|V_1|$ to 6, 3 and 1  respectively in our experiments.

{\textbf{Edges.} }
Let $E=\{E^w, E^b\}$ be the edge set, where
$E^w$ denotes the edges within each layer and $E^b$ denotes the
edges existing between different layers.
Specifically, in each intra-layer, an edge $e_{ij}\in E^w$ exists between node $v^p_i$ and $v^p_j$ if they are either
neighbour or they have common neighboring node.
For different layers, 
an edge $e^{pq}_{ij}, p\leq q$ exists between node $v^p_i$ and $v^q_j$ if  the
$i$-th part in $p$-layer involves the $j$-th part in $q$-layer.
We compute the edge weight $A^{pq}_{ij}$ for each edge as,
\begin{equation}
A^{pq}_{ij}= \exp \big(-\frac{\| x^p_i-x^q_j\|_2}{\delta} \big)
\end{equation}
where $\delta$ is a parameter.

\subsection{Graph convolutional learning}

As an extension of CNNs from regular grid to irregular graph, Graph Convolutional Networks (GCNs)~\cite{defferrard2016convolutional,kipf2016semi} have been widely studied for graph data representation and learning.
%
Our GCN learning aims to extract a contextual and compact representation for each person part  by exploring the representation information of its  neighboring parts, which thus can exploit the more discriminative information for person Re-ID.
GCN module contains several convolutional hidden layers that take a feature map matrix ${H}^{(t)}\in \mathbb{R}^{N\times d_t}$
as the input and output a feature map ${H}^{(t+1)}\in \mathbb{R}^{N\times d_{t+1}}$ by  using a graph convolution operator.
In general, we set $d_{k+1}\leq d_k$, and thus the convolution operation also provides  a kind of low-dimensional representation for graph nodes.

Let $X=[X^{(1)}\| X^{(2)}\|X^{(3)}]$ be the concatenation of $X^{(p)}$, where $X^{(p)}=(x^{p}_1, x^{p}_2 \cdots x^{p}_{n_p})$ be the extracted CNN feature collection of all parts.
Let $A$ be the whole adjacency matrix of the above hierarchical graph $G(X,A)$. We define $A$ as
$$
 A = \left(
 \begin{matrix}
   A^{11} & A^{12} & A^{13} \\
   A^{21} & A^{22} & A^{23} \\
   A^{31} & A^{32} & A^{33}
  \end{matrix}
  \right)
$$
where $A^{pq}$ is defined in Eq.(1).
Formally, given an input feature matrix $X = {H}^{(0)}\in \mathbb{R}^{N\times d_0}$ and hierarchical graph ${A}\in \mathbb{R}^{N\times N}$. Similar to GCN~\cite{kipf2016semi}, we propose to conduct the following layer-wise propagation  as
\begin{equation}
H^{(t+1)} = \sigma\big[(\epsilon \tilde{A}H^{(t)} + (1 - \epsilon)H^{(t)})\Theta^{(t)} \big]
\end{equation}
where $t = 0, 1 \cdots T-1$. $\sigma(\cdot)$ denotes an activation function. We define it as $\sigma(\cdot)=\mathrm{ReLU}(\cdot) = \max(0,\cdot)$. $\Theta = \{\Theta^{(0)}, \Theta^{(2)} \cdots \Theta^{(T-1)}\}$ denote the trainable weight matrices.
$\tilde{A}$ denotes the row-normalization of $A$~\cite{kipf2016semi}.
Parameter $\epsilon \in (0,1)$ denotes the fraction of feature information that nodes receives from their neighbors.

\subsection{Perceptron layer}

In the final perceptron layer, we combine the visual appearance information and structure information together and then adopt a FC to predict part ID label. 

For simplicity, we denote the final output feature map as $\tilde{H}=H^{(T)}=\{\tilde{h}^{(1)}_1, \tilde{h}^{(2)}_1,\tilde{h}^{(2)}_2, \tilde{h}^{(2)}_3,\tilde{h}^{(3)}_1\cdots \tilde{h}^{(3)}_6\}$.
Let $F=\{f^{(1)}_1, f^{(2)}_1, f^{(2)}_2, f^{(2)}_3,f^{(3)}_1\cdots f^{(3)}_6\}$ be the appearance feature extracted by CNN.
First, we combine $\tilde{H}$ and $F$ into $Z$ as,
\begin{equation}
z^{(p)}_i=\tilde{h}^{(p)}_i + \beta f^{(p)}_i
\end{equation}
where $\beta$ is a balancing parameter and $z^{(p)}_i$ is a part component of $Z$.
Then, for each part $z^{(p)}_i$, we adopt a FC layer to predict ID label of the corresponding person.


 \textbf{Loss function.}
For each part, we train a specific classifier by using the cross-entropy loss function $\mathcal{L}^{(p)}_i$\cite{de2005tutorial}.
The final overall loss function is designed as the aggregation of them,
\begin{equation}
\mathcal{L} = -\frac{1}{N}\sum_{p}\sum_i\mathcal{L}^{(p)}_i
\end{equation}
where $N$ is the number of parts. 
%

\begin{table*}
\caption{Comparisons results(\%) on market1501 dataset on metrics mAP, Rank-1, Rank-5 and Rank-10.}
\begin{center}
\begin{tabu} to 0.8\textwidth{X[3,l]|X[2,l]|X[l]|X[1.5,m]|X[1.5,m]|X[1.5,m]}
\hline
Methods  &reference             &mAP     &Rank-1    &Rank-5              &Rank-10\\
\hline
SVDNet\cite{Sun2017SVDNet} &ICCV2017 &62.1 &$\quad$82.3 &$\quad$92.3&$\quad$95.2\\
HydraPlus\cite{Liu_2017_ICCV} &ICCV2017 & - &$\quad$76.9 &$\quad$91.3 &$\quad$94.5\\
PAR\cite{Zhao_2017_ICCV} &ICCV2017 &63.4 &$\quad$81.0 &$\quad$92.0 &$\quad$94.7\\
PDC* \cite{su2017pose} &ICCV2017 &63.41 &$\quad$84.14 &$\quad$92.73 &$\quad$94.92\\
Rerank\cite{Zhong_2017_CVPR} &CVPR2017 &63.63 &$\quad$77.11 &$\quad$- &$\quad$-\\
MultiLoss\cite{li2017person} &IJCAI2017 &64.4 &$\quad$83.9 &$\quad$- &$\quad$-\\
SSM\cite{bai2017scalable} &CVPR2017 &68.8 &$\quad$82.21 &$\quad$- &$\quad$-\\
DPFL$^{(2+)}$ \cite{chen2017person} &CHI2017 &73.1 &$\quad$88.9 &$\quad$- &$\quad$-\\
GLAD(*)\cite{wei2017glad} &ACM MM2017 &73.9  &$\quad$89.9 &$\quad$- &$\quad$-\\
IDE+CamStyle\cite{zhong2018camera} &CVPR2018 &68.72 &$\quad$88.12 &$\quad$95.1 &$\quad$97.0\\
  $\qquad$+Re-ranking & &86.0 &$\quad$91.5 &$\quad$95.0 &$\quad$96.3\\
PSE \cite{saquib2018pose} &CVPR2018 &69.0 &$\quad$87.7 &$\quad$- &$\quad$-\\
  $\qquad$+Re-ranking & &84.0 &$\quad$90.3 &$\quad$- &$\quad$-\\
Pose-transfer\cite{Liu_2018_CVPR} &CVPR2018 &68.92 &$\quad$87.65 &$\quad$- &$\quad$-\\
BraidNet-CS+SRL\cite{wang2018person} &CVPR2018 &69.48 &$\quad$83.7 &$\quad$- &$\quad$-\\
MGCAM-Siamese\cite{song2018mask} &CVPR2018 &74.33 &$\quad$83.79 &$\quad$- &$\quad$-\\
SafeNet\cite{yuan2018safenet} &IJCAI2018 &72.7 &$\quad$90.2 &$\quad$- &$\quad$-\\
Inception-V1+OpenPose\cite{suh2018part} &ECCV2018 &76.0 &$\quad$90.2 &$\quad$96.1 &$\quad$97.4\\
  $\qquad$+Re-ranking & &89.9 &$\quad$93.4 &$\quad$96.4 &$\quad$97.4\\
HA-CNN\cite{li2018harmonious} &CVPR2018 &75.7 &$\quad$91.2 &$\quad$- &$\quad$-\\
PCB\cite{sun2018beyond} &ECCV2018 &77.3 &$\quad$92.4 &$\quad$97.0 &$\quad$97.9\\
SGGNN\cite{shen2018person} &ECCV2018 &\textbf{82.8} &$\quad$92.3 &$\quad$96.1 &$\quad$97.4\\
\hline
$\textbf{PH-GCN}$ & &79.0 &$\quad$\textbf{93.5} &$\quad$\textbf{97.4} &$\quad$\textbf{98.5} \\
$\textbf{PH-GCN+Re-ranking}$ & &\textbf{91.1} &$\quad$\textbf{94.3} &$\quad$\textbf{97.1} &$\quad$\textbf{97.8} \\
\hline
\end{tabu}
\end{center}
\end{table*}

\begin{table*}
\caption{Comparisons results(\%) on DukeMCMT-reID dataset on metrics mAP, Rank-1, Rank-5 and Rank-10.}
\begin{center}
\begin{tabu} to 0.8\textwidth{X[3,l]|X[2,l]|X[l]|X[1.5,m]|X[1.5,m]|X[1.5,m]}
\hline
Methods  &reference             &mAP     &Rank-1    &Rank-5              &Rank-10\\
\hline
BoW+kissme\cite{Zheng_2015_ICCV}&ICCV2015 &12.2 &$\quad$25.1 &$\quad$- &$\quad$-\\
LOMO+XQDA\cite{liao2015person} &CVPR2015 &17.0 &$\quad$30.8 &$\quad$- &$\quad$-\\
SVDNet\cite{Sun2017SVDNet}&ICCV2017 &56.8 &$\quad$76.7 &$\quad$- &$\quad$-\\
GAN\cite{zheng2017unlabeled} &ICCV2017 &47.1 &$\quad$67.7 &$\quad$- &$\quad$-\\
PSE \cite{saquib2018pose} &CVPR2018 &62.0 &$\quad$79.8 &$\quad$- &$\quad$-\\
$\qquad$+Re-ranking & &79.8 &$\quad$85.2 &$\quad$- &$\quad$-\\
IDE+CamStyle\cite{zhong2018camera} &CVPR2018 &53.48 &$\quad$75.27 &$\quad$84.6 &$\quad$87.9\\
$\qquad$+Re-ranking &   &73.3 &$\quad$81.4 &$\quad$88.1 &$\quad$90.8\\
Pose-transfer\cite{Liu_2018_CVPR} &CVPR2018 &56.49 &$\quad$78.52 &$\quad$- &$\quad$-\\
SafeNet\cite{yuan2018safenet} &IJCAI2018 &57.0 &$\quad$82.7 &$\quad$- &$\quad$-\\
BraidNet-CS+SRL\cite{wang2018person} &CVPR2018 &59.49  &$\quad$76.44 &$\quad$- &$\quad$-\\
Inception-V1+OpenPose\cite{suh2018part} &ECCV2018 &64.2 &$\quad$82.1 &$\quad$90.2 &$\quad$92.7\\
$\qquad$+Re-ranking &   &83.9 &$\quad$88.3 &$\quad$93.1 &$\quad$95.0\\
HA-CNN\cite{li2018harmonious} &CVPR2018 &63.8 &$\quad$80.5 &$\quad$- &$\quad$-\\
PCB \cite{sun2018beyond} &ECCV2018 &65.3  &$\quad$81.9 &$\quad$89.4 &$\quad$91.6\\
SGGNN\cite{shen2018person} &ECCV2018 &68.2 &$\quad$81.1 &$\quad$88.4 &$\quad$91.2\\

\hline
$\textbf{PH-GCN}$ & &\textbf{70.7} &$\quad$\textbf{85.0} &$\quad$\textbf{92.7} &$\quad$\textbf{94.8} \\
$\textbf{PH-GCN+Re-ranking}$ &  &\textbf{85.7} &$\quad$\textbf{88.8} &$\quad$\textbf{93.8} &$\quad$\textbf{95.9} \\
\hline
\end{tabu}
\end{center}
\end{table*}

\section{Experiments}

To verify the effectiveness of the proposed PH-GCN method, we conduct experiments on three benchmarks including  Market1501~\cite{Zheng_2015_ICCV}, DukeMTMC-reID~\cite{zheng2017unlabeled} and CUHK03~\cite{zhong2017re,li2014deepreid}.
We compare our PH-GCN with some other state-of-art methods including SSM~\cite{bai2017scalable}, IDE+CamStyle~\cite{zhong2018camera}, HA-CNN~\cite{li2018harmonious}, MGCAM-Siamese~\cite{song2018mask}, Pose-transfer~\cite{Liu_2018_CVPR}, PSE~\cite{saquib2018pose}, Inception-V1+Open-Pose~\cite{suh2018part}, SGGNN~\cite{shen2018person} and PCB~\cite{sun2018beyond}.
We implement our method with two versions, i.e., PH-GCN and PH-GCN+Re-ranking.
PH-GCN+Re-ranking further uses re-ranking~\cite{Zhong_2017_CVPR} approach to improve the learning results.

\subsection{Datasets and settings}

\noindent \textbf{Market1501}~\cite{Zheng_2015_ICCV} dataset consists of 1501 persons obtained from six camera viewpoints including five high-resolution cameras and one low-resolution camera. It contains 19,732 gallery images and 12,936 training images which are all detected by DPM~\cite{felzenszwalb2010object}.


\noindent \textbf{DukeMTMC-reID}~\cite{zheng2017unlabeled} dataset is a subset of DukeMTMC dataset~\cite{ristani2016performance}, which contains 1812 identities observed from 8 different camera viewpoints. 
It contains 1404 identities, 16522 training images, 2228 queries and 17661 gallery images, respectively.

\noindent \textbf{CUHK03}~\cite{zhong2017re,li2014deepreid} dataset contains 13164 images with 1,467 identities. Each
identity is observed from two cameras.
It contains two kinds of bounding boxes (hand-labeled, DPM-detected) and we use the former one in experiments.
 We adopt the new training/testing protocol proposed in work~\cite{zhong2017re,li2014deepreid}.

{\textbf{Evaluation metrics.}}
Following many previous works~\cite{sun2018beyond,zhong2017re}, we use Cumulative Matching Characteristic(CMC) at rank-1, rank-5 and rank-10, and the mean average precision (mAP) metrics on all datasets.
The mAP is the mean value of average precision across all queries.


\subsection{Implementation detail}
%

As shown in Figure 1, we use ResNet-50~\cite{He_2016_CVPR} pre-trained on ImageNet~\cite{russakovsky2014imagenet} as our backbone network to extract a convolutional feature map for two-stream network, respectively.
Then, we use one-layer orthodox convolution and two-layer graph convolutional network respectively to process the deeply learned part features. 
Finally, the output feature dimension in the perceptron layer is set to 256. 

We implement our model with pytorch and training the network on NVIDIA TITAN XP GPUs 12G in an end-to-end manner.
All input images are adjusted to a resolution of 384$\times$128, which is the same as that of PCB~\cite{sun2018beyond}.
Data augmentation is also adopted for training with horizontal flip, normalization.
We set the number of epochs to 60 and the batch size is set as 64 in general for all datasets.
We initialize the learning rate of backbone network to 0.1, and set the learning rate of GCN to 0.01 and other layers of network to 1.0. The learning rate is not fixed and begins to decay after 40 epochs. We train the whole network by using stochastic gradient descent (SGD)~\cite{bottou2010large} in each mini-batch.

\subsection{Comparison results}

Table 1-3 summarize the comparison results on Market1501, DukeMTMC-reID and CUHK03 datasets, respectively. The best results are marked by bold.
Overall, PH-GCN generally obtains the best results and the re-ranking operation can further improve the performance.
 More detailly, we can note that,
(1) On Market-1501 dataset, comparing with the baseline model PCB~\cite{sun2018beyond}, PH-GCN obtains rank-1 = 93.5$\%$, mAP = 79.0$\%$, which have improved performances by 1.7$\%$ and 1.1$\%$ on metrics mAP and rank-1, respectively. This clearly demonstrates the effectiveness of PH-GCN by further exploiting the structural information of parts.
PH-GCN has a higher rank value (rank-1, rank-5 and rank-10 are 1.2\%, 1.3\% and 1.1\% improved) than SGGNN~\cite{shen2018person} which employs Graph Neural Network (GNN) to guide similarity measurement for Re-ID.
(2) On DukeMTMC-reID dataset, compared with PCB~\cite{sun2018beyond}, PH-GCN results are rank-1 = 85.0$\%$ and mAP = 70.7$\%$, which also has 5.4$\%$ and 3.1$\%$ improvements on mAP and rank-1, respectively. It further demonstrates the effectiveness of our PH-GCN based representation and learning. 
Compared with SGGNN~\cite{shen2018person}, our result improves 2.5$\%$ and 3.9$\%$ on mAP and rank-1, respectively.
(3) CUHK03 dataset is known to be a very challenging dataset. PH-GCN can still improves 7.3$\%$ and 3.6$\%$ on  mAP and rank-1 when compared with PCB~\cite{sun2018beyond}. It further demonstrate the robustness of PH-GCN representation and learning.
(4) The re-ranking operation can further consistently improve the performance on all datasets.


%
%
%

\subsection{Parameter analysis}

There exist two balanced parameters $\epsilon$ (Eq.(2)) and $\beta$ (Eq.(3)) in PH-GCN.
In all experiments, we set $\{\epsilon, \beta\}$ to $\{0.75, 0.3\}$, respectively.
Empirically, the proposed model is insensitive to these parameters. When we slightly adjust the parameters, the final Re-ID results only change a little.
Figure 2 shows the results of the proposed PH-GCN with different parameters on Market1501 dataset. One can note that, when the parameters are slightly
changed, our method maintains good performance, which demonstrates the insensitivity of the proposed method w.r.t. its parameters.

\begin{table}
\caption{Accuracy comparison on CUHK03 used the new protocol.}
\begin{tabu} to 0.5\textwidth{X[2.5,l]|X[l]|X[m]|X[m]|X[m]}
\hline
Methods   &mAP      &Rank-1   &Rank-5               &Rank-10\\
\hline
BoW+kissme \cite{Zheng_2015_ICCV} &6.4 &$\quad$6.4 &$\quad$-\\
LOMO+XQDA\cite{liao2015person}  &11.5 &$\quad$12.8 &$\quad$- &$\quad$-\\
SVDNet\cite{Sun2017SVDNet} &37.3 &$\quad$41.5 &$\quad$- &$\quad$-\\
MGCAM-Siamese\cite{song2018mask}  &50.21 &$\quad$50.14 &$\quad$- &$\quad$-\\
HA-CNN\cite{li2018harmonious} &41.0 &$\quad$44.4 &$\quad$- &$\quad$-\\
PCB \cite{sun2018beyond}  &54.2 &$\quad$61.3 &$\quad$78.6 &$\quad$85.6\\

\hline
$\textbf{PH-GCN}$  &\textbf{61.5} &$\quad$\textbf{64.9} &$\quad$\textbf{81.8} &$\quad$\textbf{88.1} \\
\textbf{PH-GCN+Re-ranking} &\textbf{73.3} &$\quad$\textbf{70.6} &$\quad$\textbf{81.9} &$\quad$\textbf{87.8} \\
\hline
\end{tabu}
\end{table}

\begin{figure}[!t]
\centering
\subfloat[$\epsilon$]{\includegraphics[width=1.65in]{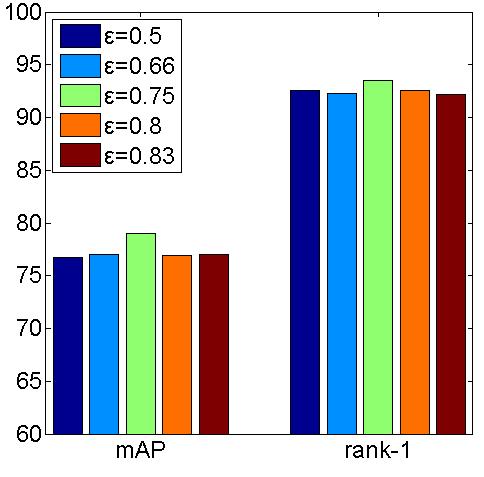}%
\label{fig_first_case}}
\hfil
\subfloat[$\beta$]{\includegraphics[width=1.65in]{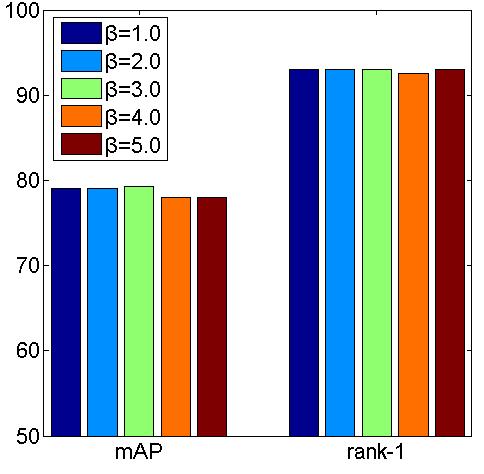}%
\label{fig_second_case}}
\caption{Results of PH-GCN with different settings of parameters $\{\epsilon, \beta\}$.}
\label{fig_sim}
\end{figure}

\begin{figure}[!t]
\centering
\subfloat{\includegraphics[width=1.7in]{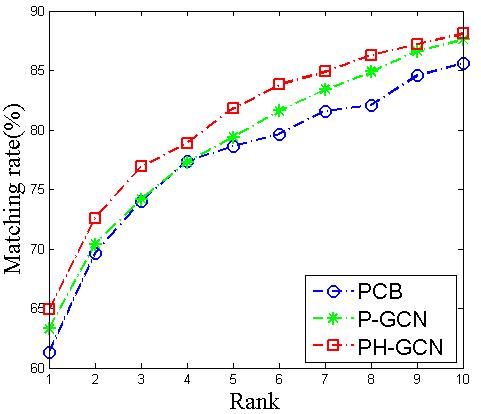}%
\label{fig_first_case}}
\hfil
\subfloat{\includegraphics[width=1.7in]{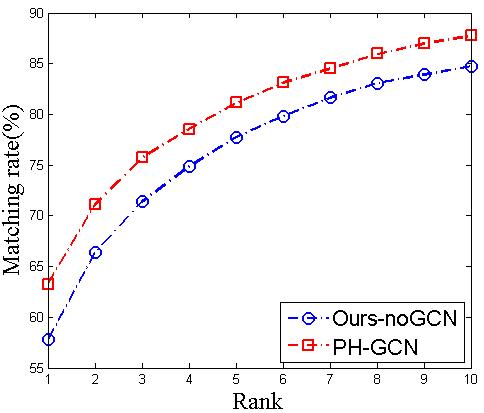}%
\label{fig_second_case}}
\caption{Performance of two variants (P-GCN, Ours-NoGCN) of the proposed Re-ID model.}
\label{fig_sim}
\end{figure}

\subsection{Ablation analysis}

In order to further understand and verify the core components (GCN module, hierarchical graph construction) of our PH-GCN model, we conduct ablation analysis experiments on CUHK03 dataset.
In particular, to verify the effectiveness of GCN module, we implement a special variant of our model, i.e., Ours-NoGCN that removes GCN module in our PH-GCN network.
To demonstrate the benefit of hierarchical graph, we implement a special variant of our model with single layer graph (denoted as P-GCN). 
 As a baseline, we also report the results of PCB~\cite{sun2018beyond}.
Figure 3 summarizes the comparison results.
Here, one can observe that
(1) Both PH-GCN and P-GCN obtain better performance than PCB, which demonstrates the effectiveness of the proposed PH-GCN (or P-GCN) by incorporating the inherent spatial relationship information among different parts.
(2) PH-GCN performs better than P-GCN, which demonstrates the benefit of hierarchical graph by capturing the structural information of both local and global cues.
(3) PH-GCN obviously outperforms Ours-NoGCN, which further shows the desired advantage of the deeply learned context-aware part representation in Re-ID task.

\section{Conclusion}


 This paper proposes a novel Part-based Hierarchical Graph Convolutional Network (PH-GCN) which aims to learn a hierarchical context-aware part feature representation for person Re-ID problem.
 %
PH-GCN also provides a general solution for object (e.g., person) representation and recognition that integrates \emph{local}, \emph{global} and \emph{structural} feature learning simultaneously in a unified end-to-end network.
Extensive experiments on three commonly used datasets demonstrate the effectiveness and benefits of the proposed PH-GCN method.

\ifCLASSOPTIONcaptionsoff
  \newpage
\fi



%



\bibliographystyle{IEEEtran}
\bibliography{ref}

\begin{thebibliography}{10}
\providecommand{\url}[1]{#1}
\csname url@samestyle\endcsname
\providecommand{\newblock}{\relax}
\providecommand{\bibinfo}[2]{#2}
\providecommand{\BIBentrySTDinterwordspacing}{\spaceskip=0pt\relax}
\providecommand{\BIBentryALTinterwordstretchfactor}{4}
\providecommand{\BIBentryALTinterwordspacing}{\spaceskip=\fontdimen2\font plus
\BIBentryALTinterwordstretchfactor\fontdimen3\font minus
  \fontdimen4\font\relax}
\providecommand{\BIBforeignlanguage}[2]{{%
\expandafter\ifx\csname l@#1\endcsname\relax
\typeout{** WARNING: IEEEtran.bst: No hyphenation pattern has been}%
\typeout{** loaded for the language `#1'. Using the pattern for}%
\typeout{** the default language instead.}%
\else
\language=\csname l@#1\endcsname
\fi
#2}}
\providecommand{\BIBdecl}{\relax}
\BIBdecl

\bibitem{Cheng2016Person}
D.~Cheng, Y.~Gong, S.~Zhou, J.~Wang, and N.~Zheng, ``Person re-identification
  by multi-channel parts-based cnn with improved triplet loss function,'' in
  \emph{Computer Vision and Pattern Recognition}, 2016.

\bibitem{si2018dual}
J.~Si, H.~Zhang, C.-G. Li, J.~Kuen, X.~Kong, A.~C. Kot, and G.~Wang, ``Dual
  attention matching network for context-aware feature sequence based person
  re-identification,'' in \emph{Proceedings of the IEEE Conference on Computer
  Vision and Pattern Recognition}, 2018, pp. 5363--5372.

\bibitem{Sun2017SVDNet}
Y.~Sun, L.~Zheng, W.~Deng, and S.~Wang, ``Svdnet for pedestrian retrieval,'' in
  \emph{IEEE International Conference on Computer Vision}, 2017.

\bibitem{shen2018deep}
Y.~Shen, H.~Li, T.~Xiao, S.~Yi, D.~Chen, and X.~Wang, ``Deep group-shuffling
  random walk for person re-identification,'' in \emph{Proceedings of the IEEE
  Conference on Computer Vision and Pattern Recognition}, 2018, pp. 2265--2274.

\bibitem{chen2018group}
D.~Chen, D.~Xu, H.~Li, N.~Sebe, and X.~Wang, ``Group consistent similarity
  learning via deep crf for person re-identification,'' in \emph{Proceedings of
  the IEEE Conference on Computer Vision and Pattern Recognition}, 2018, pp.
  8649--8658.

\bibitem{li2018harmonious}
W.~Li, X.~Zhu, and S.~Gong, ``Harmonious attention network for person
  re-identification,'' in \emph{Proceedings of the IEEE Conference on Computer
  Vision and Pattern Recognition}, 2018, pp. 2285--2294.

\bibitem{Chen_2018_CVPR}
D.~Chen, H.~Li, T.~Xiao, S.~Yi, and X.~Wang, ``Video person re-identification
  with competitive snippet-similarity aggregation and co-attentive snippet
  embedding,'' in \emph{The IEEE Conference on Computer Vision and Pattern
  Recognition (CVPR)}, June 2018.

\bibitem{ustinova2017multi}
E.~Ustinova, Y.~Ganin, and V.~Lempitsky, ``Multi-region bilinear convolutional
  neural networks for person re-identification,'' in \emph{2017 14th IEEE
  International Conference on Advanced Video and Signal Based Surveillance
  (AVSS)}.\hskip 1em plus 0.5em minus 0.4em\relax IEEE, 2017, pp. 1--6.

\bibitem{su2017pose}
C.~Su, J.~Li, S.~Zhang, J.~Xing, W.~Gao, and Q.~Tian, ``Pose-driven deep
  convolutional model for person re-identification,'' in \emph{Proceedings of
  the IEEE International Conference on Computer Vision}, 2017, pp. 3960--3969.

\bibitem{schumann2017person}
A.~Schumann and R.~Stiefelhagen, ``Person re-identification by deep learning
  attribute-complementary information,'' in \emph{Proceedings of the IEEE
  Conference on Computer Vision and Pattern Recognition Workshops}, 2017, pp.
  20--28.

\bibitem{sun2018beyond}
Y.~Sun, L.~Zheng, Y.~Yang, Q.~Tian, and S.~Wang, ``Beyond part models: Person
  retrieval with refined part pooling (and a strong convolutional baseline),''
  in \emph{Proceedings of the European Conference on Computer Vision (ECCV)},
  2018, pp. 480--496.

\bibitem{Zhao_2017_ICCV}
L.~Zhao, X.~Li, Y.~Zhuang, and J.~Wang, ``Deeply-learned part-aligned
  representations for person re-identification,'' in \emph{The IEEE
  International Conference on Computer Vision (ICCV)}, Oct 2017.

\bibitem{suh2018part}
Y.~Suh, J.~Wang, S.~Tang, T.~Mei, and K.~M. Lee, ``Part-aligned bilinear
  representations for person re-identification,'' in \emph{Computer
  Vision--ECCV 2018}.\hskip 1em plus 0.5em minus 0.4em\relax Springer, 2018,
  pp. 418--437.

\bibitem{wei2017glad}
L.~Wei, S.~Zhang, H.~Yao, W.~Gao, and Q.~Tian, ``Glad: Global-local-alignment
  descriptor for pedestrian retrieval,'' in \emph{Proceedings of the 25th ACM
  international conference on Multimedia}.\hskip 1em plus 0.5em minus
  0.4em\relax ACM, 2017, pp. 420--428.

\bibitem{zheng2018coarse}
F.~Zheng, X.~Sun, X.~Jiang, X.~Guo, Z.~Yu, and F.~Huang, ``A coarse-to-fine
  pyramidal model for person re-identification via multi-loss dynamic
  training,'' \emph{arXiv preprint arXiv:1810.12193}, 2018.

\bibitem{wang2018learning}
G.~Wang, Y.~Yuan, X.~Chen, J.~Li, and X.~Zhou, ``Learning discriminative
  features with multiple granularities for person re-identification,''
  \emph{arXiv preprint arXiv:1804.01438}, 2018.

\bibitem{bruna2013spectral}
J.~Bruna, W.~Zaremba, A.~Szlam, and Y.~LeCun, ``Spectral networks and locally
  connected networks on graphs,'' \emph{arXiv preprint arXiv:1312.6203}, 2013.

\bibitem{kipf2016semi}
T.~N. Kipf and M.~Welling, ``Semi-supervised classification with graph
  convolutional networks,'' \emph{arXiv preprint arXiv:1609.02907}, 2016.

\bibitem{niepert2016learning}
M.~Niepert, M.~Ahmed, and K.~Kutzkov, ``Learning convolutional neural networks
  for graphs,'' in \emph{International conference on machine learning}, 2016,
  pp. 2014--2023.

\bibitem{defferrard2016convolutional}
M.~Defferrard, X.~Bresson, and P.~Vandergheynst, ``Convolutional neural
  networks on graphs with fast localized spectral filtering,'' in
  \emph{Advances in neural information processing systems}, 2016, pp.
  3844--3852.

\bibitem{yan2018spatial}
S.~Yan, Y.~Xiong, and D.~Lin, ``Spatial temporal graph convolutional networks
  for skeleton-based action recognition,'' in \emph{Thirty-Second AAAI
  Conference on Artificial Intelligence}, 2018.

\bibitem{bruna2014spectral}
J.~Bruna, W.~Zaremba, A.~Szlam, and Y.~LeCun, ``Spectral networks and locally
  connected networks on graphs,'' in \emph{International Conference on Learning
  Representations}, 2014.

\bibitem{atwood2016diffusion}
J.~Atwood and D.~Towsley, ``Diffusion-convolutional neural networks,'' in
  \emph{Advances in Neural Information Processing Systems}, 2016, pp.
  1993--2001.

\bibitem{monti2017geometric}
F.~Monti, D.~Boscaini, J.~Masci, E.~Rodola, J.~Svoboda, and M.~M. Bronstein,
  ``Geometric deep learning on graphs and manifolds using mixture model cnns,''
  in \emph{IEEE Conference on Computer Vision and Pattern Recognition}, 2017,
  pp. 5423--5434.

\bibitem{velickovic2017graph}
P.~Veli{\v c}kovi{\'c}, G.~Cucurull, A.~Casanova, A.~Romero, P.~Lio, and
  Y.~Bengio, ``Graph attention networks,'' \emph{arXiv preprint
  arXiv:1710.10903}, 2017.

\bibitem{guo2018neural}
M.~Guo, E.~Chou, D.-A. Huang, S.~Song, S.~Yeung, and L.~Fei-Fei, ``Neural graph
  matching networks for fewshot 3d action recognition,'' in \emph{Proceedings
  of the European Conference on Computer Vision (ECCV)}, 2018, pp. 653--669.

\bibitem{qi20173d}
X.~Qi, R.~Liao, J.~Jia, S.~Fidler, and R.~Urtasun, ``3d graph neural networks
  for rgbd semantic segmentation,'' in \emph{Proceedings of the IEEE
  International Conference on Computer Vision}, 2017, pp. 5199--5208.

\bibitem{shen2018person}
Y.~Shen, H.~Li, S.~Yi, D.~Chen, and X.~Wang, ``Person re-identification with
  deep similarity-guided graph neural network,'' in \emph{Proceedings of the
  European Conference on Computer Vision (ECCV)}, 2018, pp. 486--504.

\bibitem{He_2016_CVPR}
K.~He, X.~Zhang, S.~Ren, and J.~Sun, ``Deep residual learning for image
  recognition,'' in \emph{The IEEE Conference on Computer Vision and Pattern
  Recognition (CVPR)}, June 2016.

\bibitem{russakovsky2014imagenet}
O.~Russakovsky, J.~Deng, H.~Su, J.~Krause, S.~Satheesh, S.~Ma, Z.~Huang,
  A.~Karpathy, A.~Khosla, M.~Bernstein \emph{et~al.}, ``Imagenet large scale
  visual recognition challenge,'' \emph{arXiv preprint arXiv:1409.0575}, 2014.

\bibitem{de2005tutorial}
P.-T. De~Boer, D.~P. Kroese, S.~Mannor, and R.~Y. Rubinstein, ``A tutorial on
  the cross-entropy method,'' \emph{Annals of operations research}, vol. 134,
  no.~1, pp. 19--67, 2005.

\bibitem{Liu_2017_ICCV}
X.~Liu, H.~Zhao, M.~Tian, L.~Sheng, J.~Shao, S.~Yi, J.~Yan, and X.~Wang,
  ``Hydraplus-net: Attentive deep features for pedestrian analysis,'' in
  \emph{The IEEE International Conference on Computer Vision (ICCV)}, Oct 2017.

\bibitem{Zhong_2017_CVPR}
Z.~Zhong, L.~Zheng, D.~Cao, and S.~Li, ``Re-ranking person re-identification
  with k-reciprocal encoding,'' in \emph{The IEEE Conference on Computer Vision
  and Pattern Recognition (CVPR)}, July 2017.

\bibitem{li2017person}
W.~Li, X.~Zhu, and S.~Gong, ``Person re-identification by deep joint learning
  of multi-loss classification,'' \emph{arXiv preprint arXiv:1705.04724}, 2017.

\bibitem{bai2017scalable}
S.~Bai, X.~Bai, and Q.~Tian, ``Scalable person re-identification on supervised
  smoothed manifold,'' in \emph{Proceedings of the IEEE Conference on Computer
  Vision and Pattern Recognition}, 2017, pp. 2530--2539.

\bibitem{chen2017person}
Y.~Chen, X.~Zhu, and S.~Gong, ``Person re-identification by deep learning
  multi-scale representations,'' in \emph{Proceedings of the IEEE International
  Conference on Computer Vision}, 2017, pp. 2590--2600.

\bibitem{zhong2018camera}
Z.~Zhong, L.~Zheng, Z.~Zheng, S.~Li, and Y.~Yang, ``Camera style adaptation for
  person re-identification,'' in \emph{Proceedings of the IEEE Conference on
  Computer Vision and Pattern Recognition}, 2018, pp. 5157--5166.

\bibitem{saquib2018pose}
M.~Saquib~Sarfraz, A.~Schumann, A.~Eberle, and R.~Stiefelhagen, ``A
  pose-sensitive embedding for person re-identification with expanded cross
  neighborhood re-ranking,'' in \emph{Proceedings of the IEEE Conference on
  Computer Vision and Pattern Recognition}, 2018, pp. 420--429.

\bibitem{Liu_2018_CVPR}
J.~Liu, B.~Ni, Y.~Yan, P.~Zhou, S.~Cheng, and J.~Hu, ``Pose transferrable
  person re-identification,'' in \emph{The IEEE Conference on Computer Vision
  and Pattern Recognition (CVPR)}, June 2018.

\bibitem{wang2018person}
Y.~Wang, Z.~Chen, F.~Wu, and G.~Wang, ``Person re-identification with cascaded
  pairwise convolutions,'' in \emph{Proceedings of the IEEE Conference on
  Computer Vision and Pattern Recognition}, 2018, pp. 1470--1478.

\bibitem{song2018mask}
C.~Song, Y.~Huang, W.~Ouyang, and L.~Wang, ``Mask-guided contrastive attention
  model for person re-identification,'' in \emph{Proceedings of the IEEE
  Conference on Computer Vision and Pattern Recognition}, 2018, pp. 1179--1188.

\bibitem{yuan2018safenet}
K.~Yuan, Q.~Zhang, C.~Huang, S.~Xiang, C.~Pan, and H.~Robotics, ``Safenet:
  Scale-normalization and anchor-based feature extraction network for person
  re-identification.'' in \emph{IJCAI}, 2018, pp. 1121--1127.

\bibitem{Zheng_2015_ICCV}
L.~Zheng, L.~Shen, L.~Tian, S.~Wang, J.~Wang, and Q.~Tian, ``Scalable person
  re-identification: A benchmark,'' in \emph{The IEEE International Conference
  on Computer Vision (ICCV)}, December 2015.

\bibitem{liao2015person}
S.~Liao, Y.~Hu, X.~Zhu, and S.~Z. Li, ``Person re-identification by local
  maximal occurrence representation and metric learning,'' in \emph{Proceedings
  of the IEEE conference on computer vision and pattern recognition}, 2015, pp.
  2197--2206.

\bibitem{zheng2017unlabeled}
Z.~Zheng, L.~Zheng, and Y.~Yang, ``Unlabeled samples generated by gan improve
  the person re-identification baseline in vitro,'' in \emph{Proceedings of the
  IEEE International Conference on Computer Vision}, 2017, pp. 3754--3762.

\bibitem{zhong2017re}
Z.~Zhong, L.~Zheng, D.~Cao, and S.~Li, ``Re-ranking person re-identification
  with k-reciprocal encoding,'' in \emph{Proceedings of the IEEE Conference on
  Computer Vision and Pattern Recognition}, 2017, pp. 1318--1327.

\bibitem{li2014deepreid}
W.~Li, R.~Zhao, T.~Xiao, and X.~Wang, ``Deepreid: Deep filter pairing neural
  network for person re-identification,'' in \emph{CVPR}, 2014.

\bibitem{felzenszwalb2010object}
P.~F. Felzenszwalb, R.~B. Girshick, D.~McAllester, and D.~Ramanan, ``Object
  detection with discriminatively trained part-based models,'' \emph{IEEE
  transactions on pattern analysis and machine intelligence}, vol.~32, no.~9,
  pp. 1627--1645, 2010.

\bibitem{ristani2016performance}
E.~Ristani, F.~Solera, R.~S. Zou, R.~Cucchiara, and C.~Tomasi, ``Performance
  measures and a data set for multi-target, multi-camera tracking,''
  \emph{arXiv preprint arXiv:1609.01775}, 2016.

\bibitem{bottou2010large}
L.~Bottou, ``Large-scale machine learning with stochastic gradient descent,''
  in \emph{Proceedings of COMPSTAT'2010}, 2010, pp. 177--186.

\end{thebibliography}

%

\end{document}